\documentclass[letterpaper, 10 pt, conference]{ieeeconf}

\IEEEoverridecommandlockouts                              

\usepackage[l2tabu,orthodox]{nag}

\usepackage[
    backend=bibtex8,
    style=ieee,
    sorting=none,
    natbib=true,
    doi=false,
    isbn=false,
    url=false,
    eprint=false,
    maxcitenames=1,
    mincitenames=1
]{biblatex}

\usepackage[pdftex,colorlinks]{hyperref}

\usepackage[printonlyused]{acronym}

\usepackage{siunitx}
\usepackage[all]{nowidow}

\usepackage[dvipsnames]{xcolor}

\usepackage{lipsum}

\usepackage{xspace} 
\newcommand{\ie}{i.e.,\xspace{}}

\usepackage[pdftex]{graphicx}

\usepackage{epstopdf}

\usepackage{import}

\usepackage{booktabs}

\usepackage{tabularx}
\usepackage{multirow, multicol}

\usepackage{amssymb,amsfonts,amsmath,amscd}

\usepackage{bm}

\newcommand{\norm}[1]{\left\Vert#1\right\Vert}
\newcommand{\abs}[1]{\left\vert#1\right\vert}

\newcommand{\bbm}{\begin{bmatrix}}
\newcommand{\ebm}{\end{bmatrix}}

\usepackage{microtype} 
\makeatletter
\usepackage[utf8]{inputenc}
\usepackage{commath}
\usepackage{mathtools}
\usepackage{mathabx}
\usepackage{siunitx}
\usepackage[ruled,vlined]{algorithm2e}
\usepackage[lofdepth,lotdepth]{subfig}
\usepackage{tikz}
\usepackage{tabularray}
\usepackage{physics}
\usepackage{bm}
\usepackage[T1]{fontenc}

\let\oldtheequation\theequation
\renewcommand\tagform@[1]{\maketag@@@{\ignorespaces#1\unskip\@@italiccorr}}
\renewcommand\theequation{(\oldtheequation)}
\makeatother

\usepackage[belowskip=0pt,aboveskip=1pt]{caption}
\usepackage[normalem]{ulem}

\newcommand{\bomega}{\bm{\omega}}

\newcommand{\bphi}{\bm{\phi}}

\newcommand{\bSigma}{\bm{\Sigma}}
\newcommand{\be}{\bm{e}}

\newcommand{\bv}{\bm{v}}

\newcommand{\R}{\mathbb{R}}

\newcommand{\bg}{\bm{g}}

\newcommand{\bq}{\bm{q}}
\newcommand{\bH}{\bm{H}}
\newcommand{\bz}{\bm{z}}

\newcommand{\bP}{\bm{P}}
\newcommand{\cL}{\mathcal{L}}
\newcommand{\cB}{\mathcal{B}_2}
\newcommand{\cS}{\mathcal{S}}
\DeclareMathOperator{\sgn}{sgn}

\DeclareSIUnit{\sqrthz}{\ensuremath{\sqrt{\text{\hertz}}}}

\usepackage{amssymb}
\usepackage{pifont}

\acrodef{ICP}{Iterative Closest Point}
\acrodef{DOF}{Degree Of Freedom}
\acrodef{INS}{Inertial Navigation System}
\acrodef{GNSS}{Global Navigation Satellite System}
\acrodef{GPS}{Global Positioning System}
\acrodef{UGV}{Uncrewed Ground Vehicle}
\acrodef{UAV}{Uncrewed Aerial Vehicle}
\acrodef{AUV}{Autonomous Underwater Vehicle}
\acrodef{MAV}{Micro Aerial Vehicle}
\acrodef{IMU}{Inertial Measurement Unit}
\acrodef{SLAM}{Simultaneous Localization and Mapping}
\acrodef{MEMS}{Micro-Electromechanical Systems}
\acrodef{GTSAM}{Georgia Tech Smoothing and Mapping Library}
\acrodef{LeGO-LOAM}{Lightweight and Ground Optimized Lidar Odometry and Mapping}
\acrodef{KF}{Kalman Filter}
\acrodef{CF}{Complementary Filter}
\acrodef{RPE}{Relative Pose Error}
\acrodef{RMSE}{Root Mean Square Error}
\acrodef{VIO}{Visual-Inertial Odometry}
\acrodef{CNN}{Convolutional Neural Network}
\acrodef{D-LIOM}{Direct Lidar-Inertial Odometry and Mapping}
\acrodef{IQR}{Interquartile range}
\acrodef{AHRS}{Attitude and Heading Reference System}

\DeclareUnicodeCharacter{0308}{¨} 

\usepackage[capitalize, nameinlink]{cleveref}
\creflabelformat{equation}{#2#1#3}
\begin{document}
\title{\LARGE \textbf{Pressure Drop: Gravity Vector Estimation from a Distributed Barometric Pressure Sensor Array}}

\author{
William Dubois,$^{1}$ Nicolas Lauzon,$^{1}$ Johann Laconte$^2$ and François Pomerleau$^{1}$
  \thanks{$^{1}$Northern Robotics Laboratory, Université Laval, Québec City, Canada,
    {\texttt{\small{$\{$william.dubois, nicolas.lauzon, francois.pomerleau$\}$ @norlab.ulaval.ca}}}}%
  \thanks{$^{2}$Université Clermont Auvergne, INRAE, UR TSCF, 63000, Clermont-
Ferrand, France {\texttt{\small{johann.laconte@inrae.fr}}}}%
}

\linepenalty=3000
\addtolength{\textfloatsep}{-0.1in}

\maketitle
\thispagestyle{empty}
\pagestyle{empty}


\begin{abstract}
While gravity is often used in mobile robotics as a global constraint for \ac{SLAM} algorithms, most estimation methods rely on indirect observation of the gravity vector.
In this paper, we propose the use of altimeter measurements to directly observe the direction of the gravity vector.
We provide a theoretical evaluation of the estimation noise and validate it through extensive experimental data, along with an evaluation of the impact of intersensor distance on our gravity vector estimates.
Experimental results demonstrate that a median error of \SI{0.87}{\degree} on the tilt estimation can be achieved with our system in semi-static scenarios.
Furthermore, we demonstrate the potential of this approach by comparing it with state-of-the-art methods based on \acp{IMU}.
We highlight that we achieve similar performance to that of \ac{IMU}-based systems in highly dynamic scenarios and even outperform in specific cases of accelerometer measurement saturation.
Finally, an additional experiment shows the potential use within a robotic system through the integration into an \ac{ICP} algorithm.
\end{abstract}



\acresetall
\section{Introduction}
In recent years, mobile robots have become increasingly popular for tasks such as search and rescue, infrastructure inspection, and environmental surveying, particularly in scenarios considered hazardous for humans.
To efficiently accomplish these missions, accurate localization and mapping are essential~\citep{Ebadi2023}.
Although \ac{SLAM} approaches can achieve strong performance~\citep{Baril2022}, they remain vulnerable to degenerate environments such as corridors, elongated forest trails, or elevators, where insufficient geometric constraints cause localization drift along under-constrained axes necessitating additional global constraints~\citep{Kubelka2022}.

\begin{figure}[t]
\centering
\includegraphics[width=\linewidth]{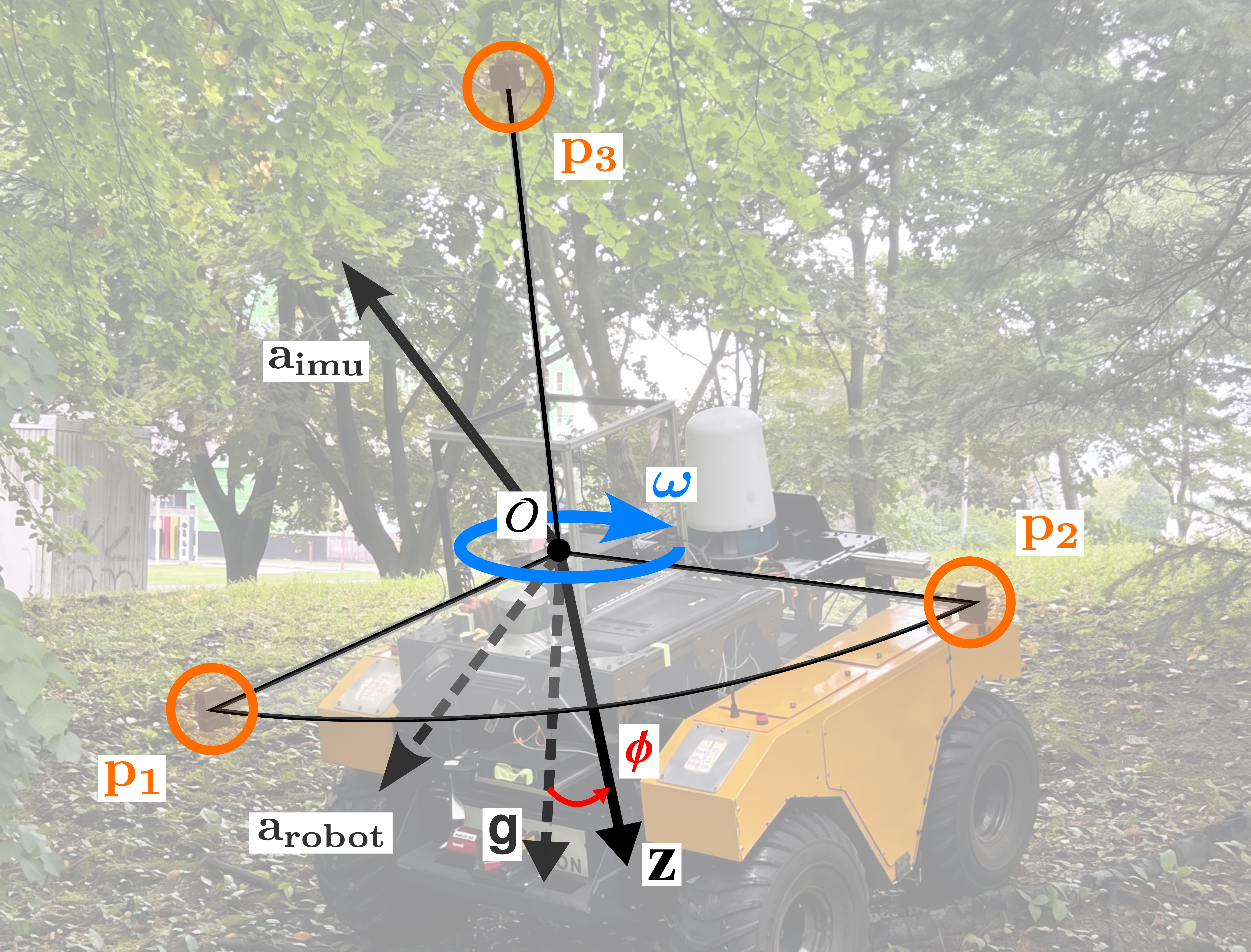}
\caption{
Field robots undergo rapid motions and forces that can challenge the \ac{IMU} for estimating gravity.
For instance, the accelerometer measurement from the \ac{IMU}, $\mathbf{a}_{\text{imu}}$ corresponds to the proper acceleration, combining the robot acceleration ($\mathbf{a}_{\text{robot}}$) and the gravity ($\bg$) reaction force.
We propose a novel way to estimate tilt ($\boldsymbol{\phi}$), the angle between the $z$-axis of the robot frame ($\mathbf{z}$) and $\bg$, via pressure sensor measurements ($p_i$) around origin $\mathcal{O}$, independent of the current motions of the robot. 
}
\label{fig:fig-intro}
\vspace{-4mm}
\end{figure}

In particular, the gravity vector provides an absolute reference for the tilt of the robotic platform and can help constrain localization and mapping solutions independently of the local geometry.
Tilt is commonly estimated from an \ac{IMU} by exploiting the gyroscope and accelerometer as indirect observations of the gravity vector~\citep{madgwick2011}.
Although this approach provides an effective and widely used solution, accelerometer measurements also contain the effects of the motion of the robot, making gravity increasingly difficult to estimate as the platform undergoes dynamic movements, as depicted in \cref{fig:fig-intro} where an \ac{UGV} climbs a hill rapidly.

Previous research has shown that barometric pressure measurements can provide accurate elevation constraints for \ac{UGV}s~\citep{Dubois2025}.
A barometric pressure measurement can be converted into an elevation value, commonly referred to as pressure elevation~\citep{Hipkin2000}.
While a single pressure sensor provides information about the elevation of its measurement point, \cref{fig:fig-intro} shows how multiple pressure sensors distributed over a known spatial configuration can provide relative elevation differences across the sensor array.
These differences can be used to estimate the gravity vector and tilt, thus inferred independently of the robot acceleration and rotational velocity measurements.
In this paper, we propose a gravity vector estimation approach based on measurements from multiple spatially distributed barometric pressure sensors.

Specifically, the main contributions of this work are as follows:
\begin{itemize}
    \item A gravity vector estimation approach that leverages measurements from multiple barometric pressure sensors, independent of linear acceleration and rotational velocity measurements;
    \item An observability and configuration analysis of an array of pressure sensors;
    \item A validation of the noise sensitivity from real measurements along with a comparison with estimates obtained from \ac{IMU} measurements;
    \item An experimental application to a state-of-the-art mapping and localization algorithm.
\end{itemize}

\section{Related work}
\label{sec:related_work}

\subsection{Tilt Estimation}
Tilt estimation typically relies on tri-axial angular velocities and linear accelerations measured by an \ac{IMU}~\citep{laidig2023vqf}.
Under static conditions, an accelerometer provides a measurement dominated by gravitational acceleration, which is a global cue on the tilt of the vehicle~\citep{agha2021nebula}.
\citet{agha2021nebula} exploited this static assumption to inject roll and pitch constraints in their \ac{SLAM} pose graph.
However, during dynamic motion, gravitational acceleration becomes indistinguishable from body accelerations~\citep{madgwick2011}.
Therefore, tilt estimation algorithms commonly combine orientation estimated by integrating gyroscope measurements and accelerometer measurements using methods such as complementary filters~\citep{Euston2008, valenti2015}, gradient-descent approaches~\citep{madgwick2011}, and Kalman filters~\citep{ahmed2017kalman}.
Robustness to dynamic motion can be further improved by explicitly modeling external accelerations~\citep{lee2012}.
Comparatively, \citet{laidig2023vqf} achieved state-of-the-art performance by first computing the orientation from the gyroscope measurements to recover an almost-inertial frame and then applying a low-pass filter to extract the gravity component of the acceleration.
Differently, \citet{Brossard2020} proposed a learning-based method to denoise \ac{IMU} gyroscope measurements and significantly reduce the integration drift.
In this paper, we propose a gravity estimation framework that allows for the recovery of the mobile platform's tilt at any given time based on singular barometric pressure measurements, relieving the necessity of prior observations and dissociating from the need for \ac{IMU} measurements.

Building on \ac{IMU}-based tilt estimation, recent works started investigating the fusion of pressure sensors into such systems.
\citet{Li2023SphericAUV} proposes to fuse depth information from a pressure sensor array along with various other sensors to improve position and orientation estimation, validating through a small-scale experimental setup with a handful of trajectories, highlighting the improvement of fusing the pressure sensor array as depth information.
Moving from depth to elevation, \citet{Dubois2025} proposed a tightly-coupled lidar-\ac{IMU}-barometer \ac{SLAM} solution, highlighting the benefits of introducing elevation constraints and achieving centimeter-level accuracy.
Similarly, \citet{Zhang2025-indoor-submeter} used differential barometric altimetry to increase vertical localization performance, achieving submeter errors along the vertical axis, which led to \SI{100}{\percent} accuracy performance in floor detection.
Likewise, \citet{Wu2026UAV} proposes a fusion algorithm improving elevation estimation for \acp{UAV} based on multiple measurements for relative elevation, both on the base station and on the mobile station, increasing the elevation accuracy in their indoor \ac{UAV} simulations.
Differently, \citet{tchonkeu2026} proposed a barometer-aided orientation estimation from \ac{IMU} measurements, showing promising performance from the added elevation information, although lacking comparison to other methods. 
In this work, we evaluate the use of elevation in tilt estimation and provide a comparison to state-of-the-art \ac{IMU}-based approaches.

Further investigation led to the estimation of tilt based on depth measurements from pressure sensors.
\citet{Zhigang2015Attitude} proposed a method to estimate the tilt of an \ac{AUV} based on four pressure sensors specifically placed on the horizontal plane of the underwater robot at rest, fusing the estimation with an \ac{INS} to increase accuracy, although only validating in simulation. 
Building upon this method, \citet{Baruch2020-AUVframework} proposed the theoretical framework to use an array of pressure sensors to compute the roll and pitch angles of an \ac{AUV}.
They provide the generic relation and equations between the depth measurements and the angles to estimate, and evaluate their method through various simulations, conducting an error analysis and an assessment of the impact of increasing the number of simulated sensors up to \num{130}.
In this work, we bring the theory further by providing an observability analysis on the mathematical foundations, showing that as little as three pressure sensors can be used to estimate the tilt of a system with the correct assumptions.
Finally, \citet{Li2023SphericAUV} also proposed including a pressure sensor array into a sensor fusion pipeline combining a depth-based tilt estimation with an \ac{IMU}-based estimation.
The authors highlight the improvement of such a fusion method through small-scale experiments, but there is a lack of evaluation of the depth-based estimation on its own.
To the authors' knowledge, elevation measurements were not investigated to estimate the orientation of an \ac{UGV}, which is the focus of this article along with an extensive evaluation of the proposed approach.

\subsection{Noise sensitivity}
\ac{MEMS} barometric pressure sensors are commonly used in the literature to compute pressure elevation, but are known to suffer from internal temperature bias and drift errors~\citep{Bolanakis2017}.
For example, \citet{Bolanakis2017} used \ac{MEMS}-based sensors in both absolute and relative elevation frameworks, and highlighted important bias sources coming from both external and internal sources.
These external sources come from changes in ambient and atmospheric conditions, for which they proposed the use of two sensors to compute relative elevation.
Secondly, the authors highlighted the importance of calibration to mitigate the internal temperature bias of \ac{MEMS} pressure sensors, and included an offset correction inherent to the \ac{MEMS} technology.
Similarly, \citet{parviainen2008differential} evaluated relative elevation estimation using pressure sensors on a car, highlighting that wind currents and ventilation have an important impact on the measurements.
Likewise, \citet{Matyja2022} modeled multiple bias sources such as non-standard sea level conditions, external disturbances, and sensor drift, managing to reduce error on barometer measurements from cycling computers.
By contrast, \citet{Sabatini2013} avoided barometric sensor biases altogether by focusing on short-term measurements, therefore ignoring the internal temperature bias and the atmospheric conditions bias.
However, in robotic deployments, the aim is to have autonomous systems able to complete missions of various lengths and complexity.
Therefore, bias mitigation based on the assumption of short-term measurements is not an effective option.
For this reason, we build on the work of \citet{Dubois2025} to mitigate the barometric pressure sensor characterization and bias, ensuring accurate barometric pressure measurements fit for elevation computation, and extend the noise sensitivity study to tilt estimation.

\section{Theory}
\label{sec:theory}
In this section, we first recall the elevation computation pipeline from \citet{Dubois2025}, before presenting the gravity vector estimation method from \citet{Baruch2020-AUVframework} based on barometric pressure sensors and extending the approach with an observability analysis, a novel special case using only three sensors, and a noise evaluation. 
Finally, the difference between our method and \ac{IMU}-based tilt estimation is highlighted by showcasing the indirect observation of the gravity vector from \ac{IMU} measurements.

\subsection{Elevation computation}
To effectively estimate tilt, the computation of elevation is necessary to, in turn, compute the relative height difference between sensors.
Specifically, we calibrated our \ac{MEMS} barometric pressure sensors using a ceramic-based reference pressure sensor.
Using these calibrated sensors, we then compute a differential relative elevation between our sensors and a chosen reference sensor on our gravity estimation setup, thus mitigating biases caused by atmospheric conditions~\citep{Bolanakis2017}.
Finally, as established by \citet{Dubois2025}, the hypsometric equation is the formulation of choice for differential elevation:
\begin{equation}
    \delta z = z_1 -z_0 = \frac{R_\text{dry}\overline{T}_v}{g} \ln{\left(\frac{p_0}{p_1}\right)},
    \label{equation_alt_hyps}
\end{equation}
where $R_\text{dry}$ is the dry air gas constant, $\overline{T}_v$ is the virtual temperature assumed to be constant, $g$ is the gravity constant and $p_0$, $p_1$ are two pressure measurements.

\subsection{Tilt estimation}

In this section, we pose the theory to estimate the gravity vector using computed elevation and extend the theory with an observability and noise analysis.
\citet{Baruch2020-AUVframework} provided the theory to estimate tilt in the local frame of the robot using a network of $N$ pressure sensors.
First, we define the set of pressure sensor coordinates in the local frame as $\left\{ \bq_0, \dots, \bq_{N-1} \right\},$ where $q_i\in\R^3$.
The pressure sensors only provide their elevation.
Therefore, in the local robot frame and taking sensor $\bq_0$ as the origin, the relative elevation of sensor $\bq_i,i\geq1$ is given by
\begin{equation}
\begin{aligned}
    \delta z_i-\delta z_0 &= \norm{\bq_i-\bq_0} \cdot \cos(\alpha)
        &= \left(\bq_i-\bq_0\right)^T\bg,
\end{aligned}
\end{equation}
where $\alpha$ is the angle between the gravity vector and the relative position  $\bq_i-\bq_0$.
Using $h_i = \delta z_i - \delta z_0$, the measurement vector is explicited as $\bz=\begin{bmatrix}h_1 & \cdots & h_{N-1}\end{bmatrix}^T\in\R^{N-1}$. 
The equation linking the gravity vector in the local frame to the measured elevations is thus
\begin{equation}
    \bz=\bP\bg,
    \label{eq:zPg}
\end{equation}
where we define the stacked quantity $\bP \in \R^{N-1\times3}$ as
\begin{equation}
\bP=
    \begin{bmatrix}
        \bq_1-\bq_0&
        \dots&
        \bq_{N-1}-\bq_0
    \end{bmatrix}^T.
    \label{eq:Pdef}
\end{equation}
\subsubsection{General case for $N\geq4$}
\Cref{eq:zPg} is solved for gravity as 
\begin{equation}
    \bg = \bH\bz, \quad\text{with}\quad \bH = (\bP^T\bP)^{-1}\bP^T .
    \label{eq:defH}
\end{equation}
This linear system has a unique solution if the rank of $\bP$ equals $3$, meaning that we need at least four non-coplanar sensors, as one of them is used as the origin. 
Note that such a condition is the same as in the work of \citet{Yoon2023_observability}.
However, one can observe that the constraint on the gravity vector norm has not been directly taken into account.
We show in the next section, in the special case of $N=3$, that the norm constraint can allow us to find a solution for the gravity vector up to a sign.

\subsubsection{Special case for $N=3$}
In this configuration, the rank of $\bP$ cannot exceed 2, and thus cannot alone find a unique solution without the addition of a constraint.
In fact, the space of solutions is a line of dimension one in the space $\R^3$ denoted $\cL$.
The addition of the norm constraint such that $\norm{\bg}=1$ adds that the solutions must lie on the L2-ball of radius 1, noted  $\cB$.
As such, the solutions are the set 
\begin{equation}
    \cS = \cL \cap \cB.
    \label{eq:set}
\end{equation}
This set can have zero, one, or two elements depending on how the line intersects the sphere.

Without loss of generality, the observation matrix $\bP$ can be written as 
\begin{equation}
\bP = \begin{bmatrix}
            a_x&0&0\\
            b_x&b_y&b_z
            \end{bmatrix},
\end{equation}
meaning that we can always find a constant frame of reference where the origin sensor and first pressure sensor are aligned with the $x$-axis.
As such, the line spanning from the linear part of \cref{eq:zPg} is
\begin{equation}
    \cL = \left\{ \begin{bmatrix}
        h_1/a_x \\
        k_1+ k_2 z \\ 
        z
    \end{bmatrix} \middle| z\in\R \right\},
\end{equation}
with 
\begin{equation}
k_1=\frac{h_2a_x-b_x h_1}{b_ya_x} \quad\text{and}\quad k_2=-\frac{b_z}{b_y}.
\end{equation}

From $\cL$, we first investigate the existence of a solution to \cref{eq:set}.
The line is contained in a plane parallel to the \textit{y-z} plane at the coordinate $x=h_1/a_x$.
From geometry, the intersection between $\cB$ and this plane is a circle of radius
\begin{equation}
    r = \sqrt{1 - \left(\frac{h_1}{a_x}\right)^2},
\end{equation}
being defined if
\begin{equation}
        \abs{h_1} \leq a_x,
        \label{eq:constraint_1}
\end{equation}
which is enforced by the mechanical constraints of the system.

The condition for a circle of radius $r$ and a line of equation $y=k_1+ k_2 z$ to intersect is determined via the discriminant $\Delta$, defined as
\begin{equation}
    \Delta = \left(1 - \left(\frac{h_1}{a_x}\right)^2\right)(1+k_2^2)-k_1^2.
    \label{eq:Delta}
\end{equation}
Therefore, $\cS$ is not empty if and only if
\begin{equation}
\begin{aligned}
    &\Delta \geq 0 \\
    \Leftrightarrow\quad &
     \frac{b_x h_1 - \gamma}{a_x}  \leq h_2 \leq \frac{b_x h_1 + \gamma}{a_x}, \\
     &\qquad\text{with}\quad\gamma=  \sqrt{\left(b_z^2 + b_y^2\right)\left(a_x^2 - h_1^2\right)}.
    \label{eq:constrainth_2}
\end{aligned}
\end{equation}

Note that the only free variable is the elevation $h_1$, as all the other quantities are defined by the mechanical structure of the setup. 
The maximal possible value of $h_2$ can be found where the gradient of both bounds vanishes, at $h_1 = \pm\frac{a_x b_x}{\norm{b}}$.
Replacing in \cref{eq:constrainth_2}, we find that
\begin{equation}
        |h_2| \leq \sqrt{b_x^2+b_y^2+b_z^2},
        \label{eq:constraint_2}
\end{equation}
meaning that the relative elevation between sensors $\bq_2$ and $\bq_0$ cannot exceed the physical distance between the two sensors.
Finally, the set of solutions is 

\begin{equation}
    \cS = \begin{cases}
        \varnothing &\text{if } |h_2|>\sqrt{b_x^2+b_y^2+b_z^2} \\
        \{v_1, v_2\} &\text{otherwise},
    \end{cases}
\end{equation}
with
\begin{equation}
    v_1, v_2 = \begin{bmatrix}
        h_1/a_x \\
        \left(k_1 \pm |k_2|\sqrt{\Delta}\right) / (1+k_2^2) \\
        \left(-k_1k_2 \pm \sgn{k_2}\sqrt{\Delta}\right) / (1+k_2^2)        
    \end{bmatrix}.
\end{equation}
The only case where $\cS$ contains a unique solution is when $\Delta=0$, meaning that one pair of pressure sensors is perfectly aligned with gravity.
Apart from this very specific case, using three pressure sensors will always lead to an ambiguity that cannot be lifted.
While a full analysis of the impact of this ambiguity is out of the scope of this paper, we may point out that this ambiguity can be mitigated with additional assumptions on the system.
In the event that the ambiguities are far enough from each other ($\Delta\gg0$), a car-like system, for instance, would be able to deduce the real tilt via a prior on acceleration, as well as simply assuming that the system cannot be upside down.

\subsubsection{Evaluation of the noise}
\label{sec:theory_plane_covariance}
In this section, we evaluate the impact of sensor positioning on the covariance of the estimated gravity vector.
From \cref{eq:zPg} we model Gaussian noise in the measurements as
\begin{equation}
    \bz = \bP\bg + \bomega, \text{ where } \bomega\sim\mathcal{N}(\bm{0},\bSigma_{\bomega}).
\end{equation}
Solving for gravity, the equation can be rewritten as
\begin{equation}
    \bg = \bH\bz + \bphi, \quad\bphi\sim\mathcal{N}(\bm{0},\bSigma_{\bphi}),
\end{equation}
with $\bSigma_{\bphi} = \bH\bSigma_{\bomega}\bH^T$ and $\bH$ defined in \cref{eq:defH}.
Assuming independant measurements from pressure sensors, $\bSigma_{\bomega}=\sigma^2\bm{I}$ and the covariance of $\bphi$ simplifies to 
\begin{equation}
    \bSigma_{\bphi} = \sigma^2(\bP^T\bP)^{-1}.
\end{equation}
In the following, we investigate the generalized variance of $\bphi$, \ie its determinant, and its impact on gravity estimation.
Since the determinant increases with the size of the Gaussian distribution, our objective is to minimize it.
With gravity being estimated only up to its direction, $\bg$ is normalized and constrained to the unit sphere. 
Under the low-noise assumption, we locally approximate the unit sphere by its tangent plane.
Without loss of generality, we choose the local frame ${\be_1,\be_2,\be_3}$ such that $\be_3=\bg$, making the tangent plane coincide with the $\be_1$-$\be_2$ plane. 
We then denote the covariance of the projected gravity estimate by $\bSigma_{\bphi}^{\text{2D}}$, corresponding to the upper $2\times2$ block of $\bSigma_{\bphi}$ in this frame.
Using the block notation, we get
\begin{equation}
\bP^T\bP =
\begin{pmatrix}
\bm{A} & \bm{b}\\
\bm{b}^T & d
\end{pmatrix},
\label{eq:block}
\end{equation}
and via the Schur complement, the determinant of the projected covariance onto the plane perpendicular to gravity is
\begin{equation}
\begin{aligned}
    \det(\bSigma_{\bphi}^\text{2D}) &= \sigma^4\det(\bm{A}-\bm{b}d^{-1}\bm{b}^T)^{-1} \\
    &= \sigma^4\frac{d}{\det(\bP^T\bP)}.
\end{aligned}
\label{eq:cov_definition}
\end{equation}
First looking at the denominator, we have
\begin{equation}
\begin{aligned}
\det(\bP^T\bP)
=
\det\left(\sum_{i=1}^{N-1} \bv_i\bv_i^T\right)
=
\sum_{i<j<k}
\left(\bv_i\cdot(\bv_j\times\bv_k)\right)^2,
\end{aligned}
\label{eq:PtPdef}
\end{equation}
with $\bv_i=\bq_i-\bq_0$.
As such, the denominator grows as a function of the volume spanned by the vectors $\bv_i$, representing the directions of the sensors relative to $\bq_0$. For $N=4$, the denominator is maximized if $\bv_1, \bv_2$ and $\bv_3$ are perpendicular to each other. 

Regarding the numerator of \cref{eq:cov_definition}, from \cref{eq:block,eq:Pdef} we have
\begin{equation}
d = \sum_i (v_i^z)^2,
\label{eq:numerator}
\end{equation}
where $v_i^z$ denotes the third component of $\bv_i$, that is along gravity. 
The numerator therefore increases with the extent of the sensor configuration along the gravity direction and thus the increase of intersensor distances if placed along that direction.
Therefore, minimizing the projected covariance requires balancing these two effects: limiting the sensor displacement along gravity while maintaining a large three-dimensional spread. 
The corresponding optimal configuration is left for future work, while experimental evidence supporting the above analysis is provided in the results.

\subsection{Indirect observation from \ac{IMU} measurements}
First, the gravity vector can be isolated from \ac{IMU} accelerometer measurements as 
\begin{equation}
    \bg = {}^{\mathcal{G}}_{\mathcal{B}}\mathbf{R}({}^{\mathcal{B}}\tilde{\mathbf{a}}(t) - {}^{\mathcal{B}}\mathbf{a}(t) + \boldsymbol{\eta}^a(t) + \mathbf{b}^a(t)),
\end{equation}
where $\mathcal{B}$ denotes the body frame, $\mathcal{G}$ is the global gravity-aligned frame, ${}^{\mathcal{B}}\tilde{\mathbf{a}}$ is the \ac{IMU} accelerometer measurement, ${}^{\mathcal{B}}\mathbf{a}$ is the body acceleration and $\boldsymbol{\eta}$ and $\mathbf{b}$ are noise and bias affecting the measurement.
In a static situation, the body acceleration is zero and gravity can effectively be estimated from \ac{IMU} accelerometer measurements.
However, in dynamic situations other forces will be applied on the body frame, leading to the necessity of estimating the body acceleration to effectively estimate gravity.
To accomplish this estimation, a common assumption is that gravity will dominate the accelerometer signal and that partial observability is enough to do long-term tilt estimation~\citep{madgwick2011, Euston2008, valenti2015}.
Differently, other methods opt to leverage additional sensor modalities to estimate body accelerations, thus decoupling them from gravity~\citep{noh2026garlileo}.

\begin{figure}[htbp]
    \centering
    \includegraphics[width=\linewidth]{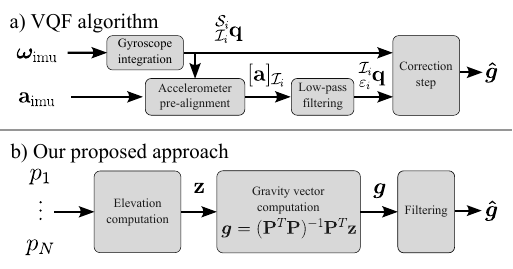}
    \caption{Comparison of the processing steps and complexity of (a) the VQF algorithm and (b) our proposed approach. Highlighting the differing inputs $\boldsymbol{\omega}_{\text{imu}}$, $\mathbf{a}_{\text{imu}}$ and pressure measurements $p_i$ and the differing transition states in each algorithm.}
    \label{fig:pipelines}
    \vspace{-2mm}
\end{figure}

In this work, the VQF algorithm~\citep{laidig2023vqf} is one of the algorithms used for comparison in the results.
\Cref{fig:pipelines} a) illustrates how it handles the partial observability of gravity.
First, the gyroscope measurement is integrated to obtain an almost-inertial frame $\mathcal{I}$ through ${}^{\mathcal{S}_i}_{\mathcal{I}_i}\mathbf{q}$, then the accelerometer measurement is pre-aligned to that frame using the quaternion to obtain an almost-inertial accelerometer measurement $[\mathbf{a}]_{\mathcal{I}_i}$.
This aligned measurement is then filtered using a hand-tuned cutoff frequency to recover an accelerometer correction quaternion ${}^{\mathcal{I}_i}_{\varepsilon_i}\mathbf{q}$ used in the correction step to finalize the tilt estimation through quaternion multiplication with ${}^{\mathcal{S}_i}_{\mathcal{I}_i}\mathbf{q}$.
In contrast, \cref{fig:pipelines} b) illustrates our proposed approach, which builds on the elevation and tilt computation steps described previously by adding a filtering step to produce the final gravity vector estimate.
This highlights the direct nature of our observation method and its lower computational complexity.

\section{Results}
\label{sec:results}
This section first details the experimental setup used to acquire data to test and validate the proposed approach.
Secondly, an evaluation of the approach is presented, highlighting the impact of the number of pressure sensors and their positions, followed by an extensive analysis of the impact of the intersensor distance on our gravity estimation.
Then, our pressure-based approach is compared to \ac{IMU}-based approaches in a highly dynamic scenario.
Finally, a qualitative evaluation of the proposed solution is shown through its use within an \ac{ICP} algorithm used for localization and mapping.

\subsection{Experimental setup}
The experimental setup consists of a custom-made gravity estimation rig, a Vicon Tracker motion capture system, and an \ac{UGV}.
The gravity estimation rig is a platform containing eight DPS310 barometric pressure sensors, a Setra model 278 industrial pressure sensor, used as a reference for sensor calibration, an Xsens MTi-10 \ac{IMU}, and a CHR-UM6 \ac{IMU}.
Two different \acp{IMU} are mounted on our system to highlight the impact of different levels of measurement noise and different levels of saturation.
\cref{table:imus} presents important specifications for both \acp{IMU}, where we can see that both sensors have a gyroscope with a saturation level of $\pm$\SI[per-mode=symbol]{2000}{\degree\per\second}, with the MTi-10 having \SI{75}{\percent} lower noise.
Moreover, the MTi-10 has a saturation level eight times higher than that of the CHR-UM6 on the accelerometer measurements, leading to highly differing performances under dynamic scenarios.
Both the CHR-UM6 \ac{IMU} and the DPS310 barometric pressure sensors are considered low-cost sensors, while the Xsens MTi-10 and the Setra model 278 are higher tier.
The gravity estimation rig was deployed in a Vicon Tracker motion capture system to accurately evaluate pressure-based gravity estimates and compare them fairly to \ac{IMU}-based estimates using a precise reference.
Finally, as shown in \cref{fig:fig-intro}, the experimental setup was installed on an \ac{UGV} to evaluate the feasibility of using our approach on a mobile robot for localization and mapping.

\begin{table}[htbp]
\centering
\caption{Major specifications for both \acp{IMU} installed on the gravity estimation rig, impacting their tilt estimation capabilities.}
\begin{tblr}{Xcc} 
\cline[1pt, black]{-}
 & \textbf{Mti-10} & \textbf{CHR-UM6} \\
\cline[1pt, black]{-}
Gyroscope measurement range (\unit[per-mode=symbol]{\degree\per\second}) & $\pm2000$ & $\pm2000$ \\
Gyroscope noise density (\unit[per-mode=symbol]{\degree\per\second\per$\sqrt{\hertz}$}) & $0.007$ & $0.03$ \\
Accelerometer measurement range (\unit{\g}) & $\pm16.0$ & $\pm2.0$ \\
\cline[1pt, black]{-}
\end{tblr}
\label{table:imus}
\end{table}

\subsection{Tilt estimation performance}

\begin{figure}[htbp]
	\centering
	\includegraphics[width=\linewidth]{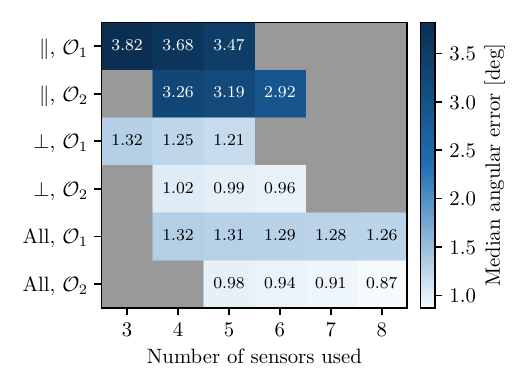}
	\caption{
    Heatmap showing the impact of the number of sensors and the planes on which they lie on tilt estimation.
    The marker $\|$ corresponds to planes parallel to gravity, $\perp$ corresponds to the plane orthogonal to gravity, and \emph{All} corresponds to the use of both parallel and orthogonal planes to gravity.
    The markers $\mathcal{O}_1$ and $\mathcal{O}_2$ respectively correspond to the use of a single sensor as the origin or the averaged measurements of two sensors for the origin.
    Gray spaces correspond to impossible combinations of sensor, planes, and origin.}
	\label{fig:perf_analysis}
    \vspace{-2mm}
\end{figure}

Here, we present the results of the proposed pressure-based gravity estimation pipeline, along with an analysis of the impact of the number of sensors used, the spatial configuration, and the sensitivity to measurement noise.
For this evaluation, the gravity estimation rig used an intersensor distance of \SI{190}{\centi\metre}, which was the longest possible configuration for our system.
In total, four hours of experimental data were recorded over \num{13} experiments.
\Cref{fig:perf_analysis} presents the obtained results with rows separating the different planes created by the sensors used for the gravity estimation.
The first two rows are parallel to the gravity direction at rest ($\|$), the third and fourth rows are orthogonal to the gravity at rest ($\perp$), and the last two rows are a combination of sensors lying on both parallel and orthogonal planes with regard to the gravity (All).
Furthermore, $\mathcal{O}_1$ and $\mathcal{O}_2$ respectively represent the use of a single sensor for the origin or the averaged measurements of two sensors.
Finally, gray spaces correspond to impossible combinations of sensors, planes, and origin.

The rows marked with $\|$ and $\perp$ show that the experimental data is in agreement with the theoretical noise analysis, proving that having sensors on an orthogonal plane to gravity will clearly outperform having them on a plane parallel to gravity by lowering the covariance numerator.
As a practical note, such an orthogonal plane with regard to gravity is easily accessible for most vehicles (e.g., the roof), leading to an easier sensor positioning than computing an optimal combination of positions.
Furthermore, we highlight that for any given choice of sensor placement, increasing the number of sensors used will lead to a decrease in the angular error of the gravity estimate.
Additionally, comparing $\mathcal{O}_1$ and $\mathcal{O}_2$ rows, a significant increase in performance is visible for an averaged origin measurement, of up to \SI{28}{\percent}, for any given number of sensors and any sensor placement.
This increase originates in the averaging of two origin sensor measurements together, reducing the noise level of the reference measurement, which is then used in combination with all other sensors, leading to reduced noise on all computed elevations.
Overall, in accordance with the theoretical noise analysis, we see that the positioning of sensors as well as the measurement noise on the origin have a significant impact on the accuracy of the tilt estimation.

Finally, going from an orthogonal plane to using all planes without additional sensors can degrade the quality of the estimation, as can be seen when comparing the $\perp$ rows to the All rows.
In fact, in accordance with the theory, we increase the covariance when expanding along the gravity direction; averaging the origin or additional sensors becomes a necessity to then increase the accuracy of the estimation.
For the remainder of this work, we opt to use the best combination available on our experimental setup, which corresponds to using all eight of our pressure sensors along with an averaged origin.
\subsection{Intersensor distance analysis}
To enable further evaluation of our tilt estimation system, we investigate the impact of intersensor distance on the angular error with the intent of having a reasonable system size to manipulate and install on a robotic platform.
This evaluation was conducted with two experiments at every evaluated distance, both combining static and dynamic measurements, with the majority of the experiment being static.
The dynamic part of the experiments consisted of realistic and fluid movements that could be induced on the system by a moving robot or vehicle.
This set of experiments was conducted with intersensor distances ranging from \SI{60}{\centi\metre} to \SI{190}{\centi\metre}, where the limits were chosen based on the mechanical limitations of the estimation rig and limitations from the Vicon Tracker system.
While our method is generic and accommodates individual distances between sensors, we keep the intersensor distances equal for all directions of our experimental setup for ease of manipulation and comparison.

\begin{figure}[htbp]
\vspace{-3mm}
	\centering
	\includegraphics[trim={13px 0 0 0}, width=\linewidth]{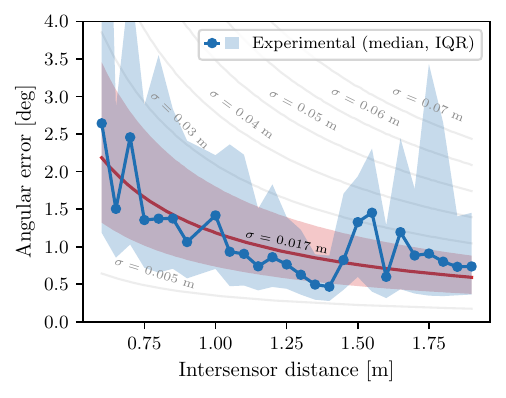}
\caption{Impact of intersensor distance on the angular error in degrees, computed by comparing our pressure-based approach to the Vicon Tracker system ground truth.
Experimental data (blue) are compared against theoretical data (red, \SI{1.7}{\centi\metre} pressure measurement standard deviation) and background curves showing expected error for different noise levels.}
	\label{fig:distance_analysis}
    \vspace{-2mm}
\end{figure}

The impact of intersensor distance is presented in \cref{fig:distance_analysis}, where the theoretical standard deviation on pressure measurement corresponds to $\sigma$ in \cref{eq:cov_definition}.
The theoretical data were generated using Monte Carlo simulations, which sampled measurements in the given Gaussian distributions for every centimeter from \SI{60}{\centi\metre} to \SI{190}{\centi\metre} to match the experimental distances, and the error was computed from the estimated gravity using sampled measurements and the expected gravity from exact distance values.
The background curves further illustrate expected angular error trends across different levels of pressure measurement noise.

First, we highlight that the experimental results match the theoretical data: the median values align with the theoretical median for most distances and remain within the \ac{IQR} across most of the curve.
With future increases in technology, even lower error could be attainable, with the lowest background curve showing what to expect from sensors with a standard deviation of \SI{0.5}{\centi\metre}.
In our case, a discrepancy between theoretical data and experimental data is visible at higher distances, where the experimental data has an increase in error around an intersensor distance of \SI{1.5}{\m} and hits a plateau.
This evolution of the error can be explained by the experimental setup used to conduct the experiments. 
To ease the movements, the weight of the tilt estimation rig was limited, leading to a limitation in cross-section dimension for the aluminum extrusions used to build the rig. 
This limitation led to increased flexibility at longer distances, thus increasing the error between the pressure-based estimation and the ground truth by breaking the rigid body assumption.
Finally, the reduction in error slows after \SI{1.25}{\metre}, making this distance an acceptable trade-off between error and size for our vehicle.
We therefore limited our experimental setup to this intersensor distance, which eased further manipulation and, ultimately, integration with our mobile robot platform.

\subsection{Highly dynamic scenario evaluation}
\begin{figure}[htbp]
	\centering
	\includegraphics[width=\linewidth]{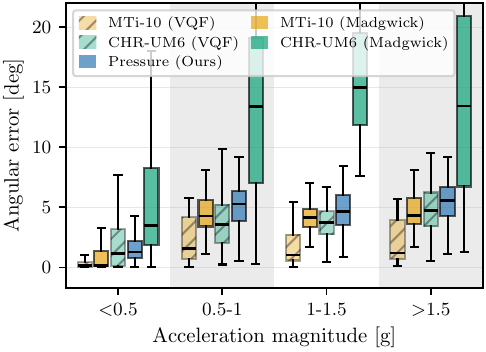}
	\caption{Impact of the acceleration magnitude on the tilt estimation angular error in highly dynamic scenarios.}
	\label{fig:imu-dynamic-case}
    \vspace{-2mm}
\end{figure}

This section presents a comparison between the performances of our gravity estimation pipeline and that of both the VQF and Madgwick filter algorithms, using the recommended default parameters, on the two different \acp{IMU} available on our experimental setup.
To evaluate the relevance of our approach, we compare the solutions through a highly dynamic scenario in which we generate high accelerations while creating tilt changes.
\Cref{fig:imu-dynamic-case} shows the results obtained with each approach during this experiment.
The data are divided into four clusters based on acceleration magnitude, with boundaries set according to the CHR-UM6 \ac{IMU}'s $\pm$\SI{2}{\g} accelerometer measurement range. 
First, we highlight that our proposed estimation method, although impacted by the acceleration, reaches consistent performance across all higher acceleration clusters, rather than increasing with every cluster. 
By contrast, the error for both \acp{IMU} using the VQF algorithm and for the MTi-10 using the Madgwick filter shows a slight upward trend with the increase of acceleration magnitude.
Additionally, the results show that the Madgwick filter gives more confidence to the accelerometer than the VQF algorithm, with the CHR-UM6 \ac{IMU} having a significant increase in error at higher accelerations, coming from saturated accelerometer measurements, leading to completely erroneous tilt estimates.
Overall, our pressure-based approach mitigates the impact of high accelerations on tilt estimation as the error stays constant with higher acceleration levels.
It also achieves similar results to the compared \ac{IMU}-based algorithms in such highly dynamic scenarios, and outperforms the widely-used Madgwick filter when the accelerometer measurements are saturated, as depicted through the use of the CHR-UM6 \ac{IMU}.

\subsection{Feasibility in mapping algorithms}
\begin{figure}[htbp]
    \centering
    \includegraphics[width=\linewidth]{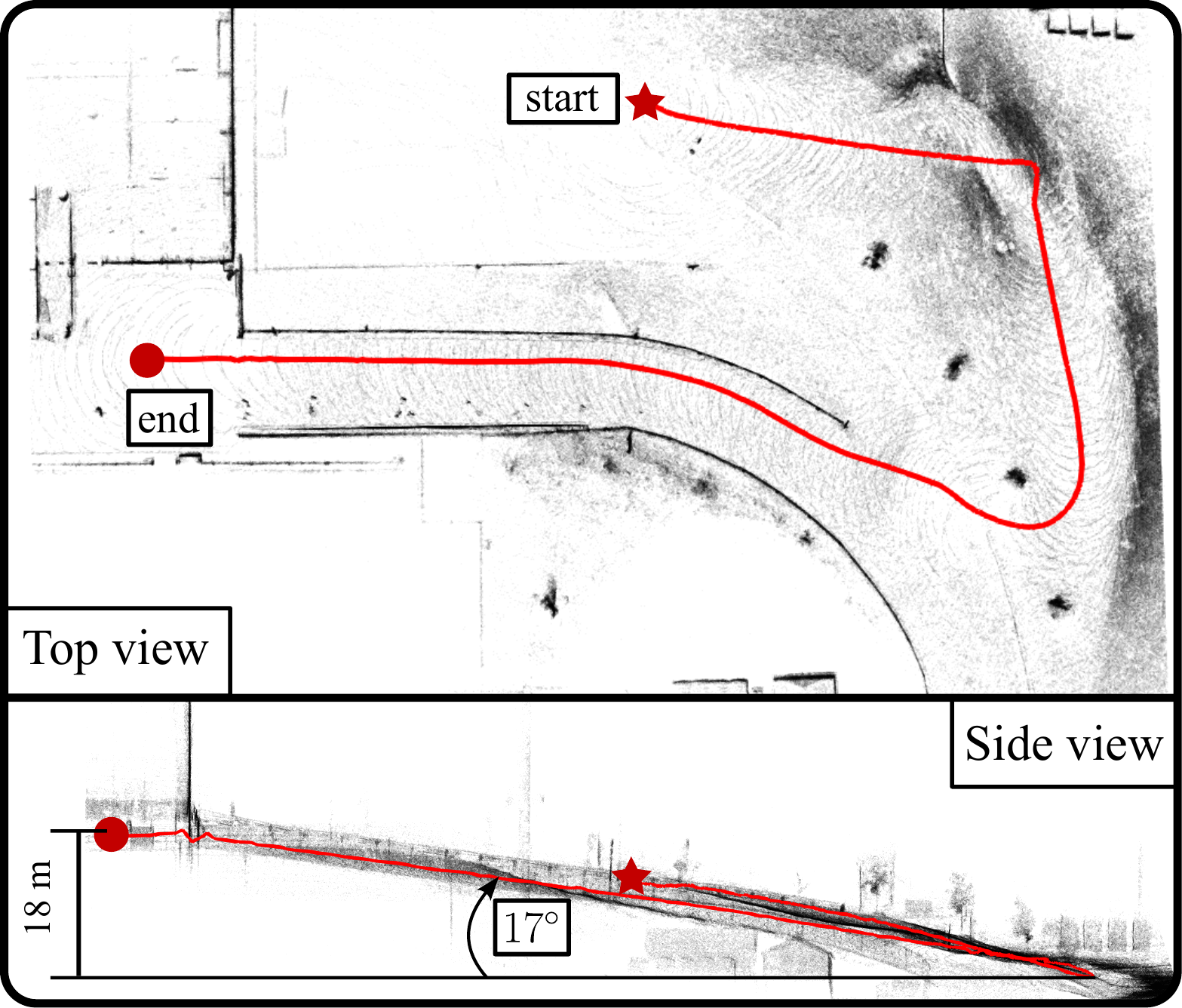}
    \caption{Qualitative result depicting the map (black) and trajectory (red) computed from an outdoor experiment conducted using the proposed pressure-based gravity vector estimation approach as part of the odometry prior for an \ac{ICP} localization and mapping algorithm.}
    \vspace{-3mm}
    \label{fig:icp_map}
\end{figure}

Finally, we highlight the feasibility of integrating our gravity estimation approach into a state-of-the-art localization and mapping algorithm.
The integration of our estimate comes as an odometry used to pre-align point clouds, with the final localization being refined using the \ac{ICP} algorithm.
\Cref{fig:icp_map} presents the top and side views of the resulting map and trajectory (red) of the experiment conducted with the robotic system depicted in \cref{fig:fig-intro}.
The figure qualitatively shows that the computed odometry from our pressure-based gravity vector is satisfactory to produce high-quality maps through experiments involving tilt evolution such as going down and up hills while maintaining structural integrity of the surrounding environment and obstacles such as trees, buildings, and fences.
The conducted experiment, although purposely exciting the tilt, is limited to a rather constrained environment and a trajectory only spanning \SI{110}{\meter} with \SI{30}{\meter} of total vertical displacement on a \SI{17}{\degree} incline slope.
Nevertheless, this experiment provides the first demonstration of the configuration's viability in an outdoor setting.

\vspace{-1mm}
\section{Conclusion}
\vspace{-1mm}
\label{sec:conclusion}

In this work, we proposed a gravity estimation pipeline leveraging barometer measurements to directly observe the gravity direction.
We first brought forward the equations enabling that direct observation, showing that four sensors or more are necessary to have a fully observable system, but also that with the right assumptions it can be done with only three sensors.
We evaluated the impact of pressure sensor positioning and intersensor distance on the error, and validated our theoretical contributions through extensive experimental validation.
Furthermore, we compared our proposed approach to \ac{IMU}-based estimations in highly dynamic scenarios, showing that our approach is uncorrelated to typical \ac{IMU}-based estimation problems such as high accelerations, offering a redundancy option for tilt estimation.
Finally, we integrated our estimation approach in a state-of-the-art mapping and localization algorithm to qualitatively show the functionality of our approach.

Future work includes an investigation of the impact of vehicle speed on the estimation, and the integration of the gravity estimation pipeline in a complete \ac{SLAM} algorithm along with its uncertainty.
Furthermore, fusion of \ac{IMU} measurements along with pressure-based gravity estimation could provide increased performance combining direct and indirect observations of the gravity vector along with global constraint redundancy, which could be used for fault detection towards safe autonomous robotics.
Finally, now that the solution was validated, more field experiments are required to investigate the impact of weather conditions, sustained vibrations, and vehicle speed.

\vspace{-1mm}
\section*{Acknowledgment}
\vspace{-1mm}
This work was supported by the  Natural  Sciences and Engineering  Research  Council of  Canada  (NSERC)  through grants ALLRP 585289-23 and ES D 589528 and by the Canada Research Chair on Field Robotics (CRC-2023-00396).
Generative AI (Claude, Anthropic) was used to assist in generating the plotting code for the data-driven result figures; the underlying data, analysis, and figure content were recorded, produced, and verified by the authors.
\vspace{-1mm}
\renewcommand*{\bibfont}{\footnotesize}

\IEEEtriggeratref{6}
\IEEEtriggercmd{\enlargethispage{-0.1in}}

\printbibliography
\end{document}